\documentclass{article}

\usepackage{wrapfig}

\usepackage[dblblindworkshop, final]{neurips_2025}
\usepackage[utf8]{inputenc} % allow utf-8 input
\usepackage[T1]{fontenc}    % use 8-bit T1 fonts
\usepackage{hyperref}       % hyperlinks
\usepackage{url}            % simple URL typesetting
\usepackage{booktabs}       % professional-quality tables
\usepackage{amsfonts}       % blackboard math symbols
\usepackage{nicefrac}       % compact symbols for 1/2, etc.
\usepackage{microtype}      % microtypography
\usepackage{xcolor}         % colors
\usepackage{graphicx}
\usepackage{subcaption}
\usepackage{amsmath}

\title{Saliency Guided Longitudinal Medical Visual Question Answering}

\author{%
  Jialin Wu\\
  Dept. of Computer Science and Engineering\\
  University of California, San Diego\\
  San Diego, CA 92037 \\
  \texttt{jlwu@ucsd.edu} \\
  \And
  Xiaofeng Liu \\
  Dept. of Radiology and Biomedical Imaging \\
  Yale University \\
  New Haven, CT 06510 \\
  \texttt{xiaofeng.liu@yale.edu} \\
}

\begin{document}
\maketitle

\begin{abstract}
Longitudinal medical visual question answering (Diff-VQA) requires comparing paired studies from different time points and answering questions about clinically meaningful changes. In this setting, the difference signal and the consistency of visual focus across time are more informative than absolute single-image findings. We propose a saliency-guided encoder–decoder for chest X-ray Diff-VQA that turns post-hoc saliency into actionable supervision. The model first performs a lightweight near-identity affine pre-alignment to reduce nuisance motion between visits. It then executes a within-epoch two-step loop: step 1 extracts a medically relevant keyword from the answer and generates keyword-conditioned Grad-CAM on both images to obtain disease-focused saliency; step 2 applies the shared saliency mask to both time points and generates the final answer. This closes the language–vision loop so that the terms that matter also guide where the model looks, enforcing spatially consistent attention on corresponding anatomy. On Medical-Diff-VQA, the approach attains good performance on BLEU, ROUGE-L, CIDEr, and METEOR while providing intrinsic interpretability. Notably, the backbone and decoder are general-domain pretrained without radiology-specific pretraining, highlighting practicality and transferability. These results support saliency-conditioned generation with mild pre-alignment as a principled framework for longitudinal reasoning in medical VQA.
\end{abstract}

% ; and  an selection criterion that fuses CIDEr with METEOR to ensure medical keyword usage and semantic adequacy.

\section{Introduction}
\label{sec:intro}
Medical Visual Question Answering (VQA) aims to answer open-ended clinical questions based on medical images, serving as a critical bridge from visual perception to clinical decision support ~\citep{survey}. Numerous medical VQA approaches in recent years have relied on pretrained visual or multimodal models~\citep{GLoRIA, llava, pmcvqa}. However, most of these works focus on a single time-point following the natural image VQA tasks. 
Radiologists routinely compare current and prior studies to localize change, judge progression, and reconcile apparent discrepancies. 

Difference/longitudinal visual question answering (Diff-VQA) operationalizes this workflow by conditioning answers on paired images acquired at two time points, where the difference is often the signal of interest rather than absolute appearance \citep{ekaid}. Recent benchmarks and methods for longitudinal chest X-rays have made this task concrete by supplying paired images, questions, and change-focused references~\citep{mimic-diff-2, dataset1, dataset2}. Building on these resources, several approaches adapt vision–language models or design task-specific architectures to better capture temporal discrepancies, including prior work that emphasizes longitudinal pretraining \citep{plural}, residual alignment in the feature or pixel space \citep{reai}, or region-level retrieval and mixing \citep{regio}.
However, their attention in different time-points is not explicitly encouraged to be consistent, which is essential for the compare and contrast to explore the difference.

Saliency maps are a type of saliency visualization used to interpret deep learning models. In medical imaging tasks, they are widely employed to present verifiable evidence to clinicians and enhance model interpretability and trustworthiness~\citep{sal-driven}. In medical VQA tasks, researchers frequently employ attention/saliency visualizations to verify whether models focus on relevant image evidence when generating responses, reflecting the critical need for explainable and traceable reasoning processes in high-stakes medical contexts~\citep{survey}. However, existing medical-VQA models often treat saliency as a post-hoc explanation~\citep{medsal1:jin2021mapdoesfitall, medsal2, medsal3} rather than incorporating it as intrinsic supervision during training. In longitudinal settings this is a missed opportunity because consistent focus on corresponding anatomy across the two time points is essential to answering difference-type questions faithfully.

We introduce a saliency-guided longitudinal VQA framework that makes saliency actionable during learning. The method has two design principles: (i) make the two images geometrically comparable and (ii) ensure that what the model says it cares about also determines where it looks at both time points, inspired by natural image co-attention~\citep{co-atten,co-sal}. Specifically, we have two modules. $\bullet$ \textbf{Micro pre-alignment.} A lightweight CNN-based module applies a near-identity affine warp to the current study to mitigate small pose and scale variations without overfitting or erasing true changes \citep{stn}. $\bullet$ \textbf{Keyword-conditioned shared saliency.} Within each training epoch we run a two-step loop akin to the Expectation Maximization Algorithm~\citep{em}. First, a language model extracts one clinically salient keyword from the ground-truth answer. We compute keyword-conditioned Grad-CAM on both time points and take their union to form a shared saliency mask. Second, we re-encode masked images and generate the answer with a multimodal decoder, which ties linguistic supervision to spatial evidence and encourages consistent attention on the same anatomical regions across time.

The main contributions can be summarized as:

$\bullet$We formalize a simple and effective way to enforce \emph{spatially consistent attention} across paired images by using keyword-conditioned, shared saliency as a training signal for Diff-VQA.

$\bullet$We couple this with a minimal pre-alignment module that improves longitudinal comparability while preserving true differences.

$\bullet$We demonstrate competitive results on Medical-Diff-VQA using only general-domain pretrained backbones and decoders, yielding strong practicality and inherent interpretability without radiology-specific pretraining.

% one category fine-tunes CNN/ViT backbones trained on ImageNet; another performs image-text pre-training in radiology using image-report pairs; and approaches transferring VLM models fine-tuned on general tasks to biomedical imaging. 

% Model consists of a STN registration module~\citep{stn}, an.
% \begin{figure}[!htbp] 
% \centering 
% \includegraphics[width=0.7\linewidth]{arch.pdf} \vspace{\baselineskip} 
% \caption{The overview of our proposed method. The saliency extraction and applicaiton are based on the after-registration images.}  
% \label{fig:arch} 
% \end{figure}

\section{Methods}
\label{sec:meth}
 
We use the longitudinal chest radiograph Diff-VQA dataset, Medical-Diff-VQA~\citep{ekaid,mimic-diff-2}, which constructs samples from paired studies of the same subject at two time points together with a difference-focused question–answer pair. The dataset is derived from MIMIC-CXR~\citep{MIMIC-CXR} and MIMIC-CXR-JPG~\citep{mimic-cxr-jpg} and was obtained from PhysioNet~\citep{physionet}. All usage follows the PhysioNet credentialed-access license and de-identification guidelines. The corpus contains 164{,}223 samples, split into 131{,}556 for training, 16{,}278 for validation, and 16{,}389 for testing.

\begin{figure}[!htbp] 
  \centering
  \includegraphics[width=\linewidth]{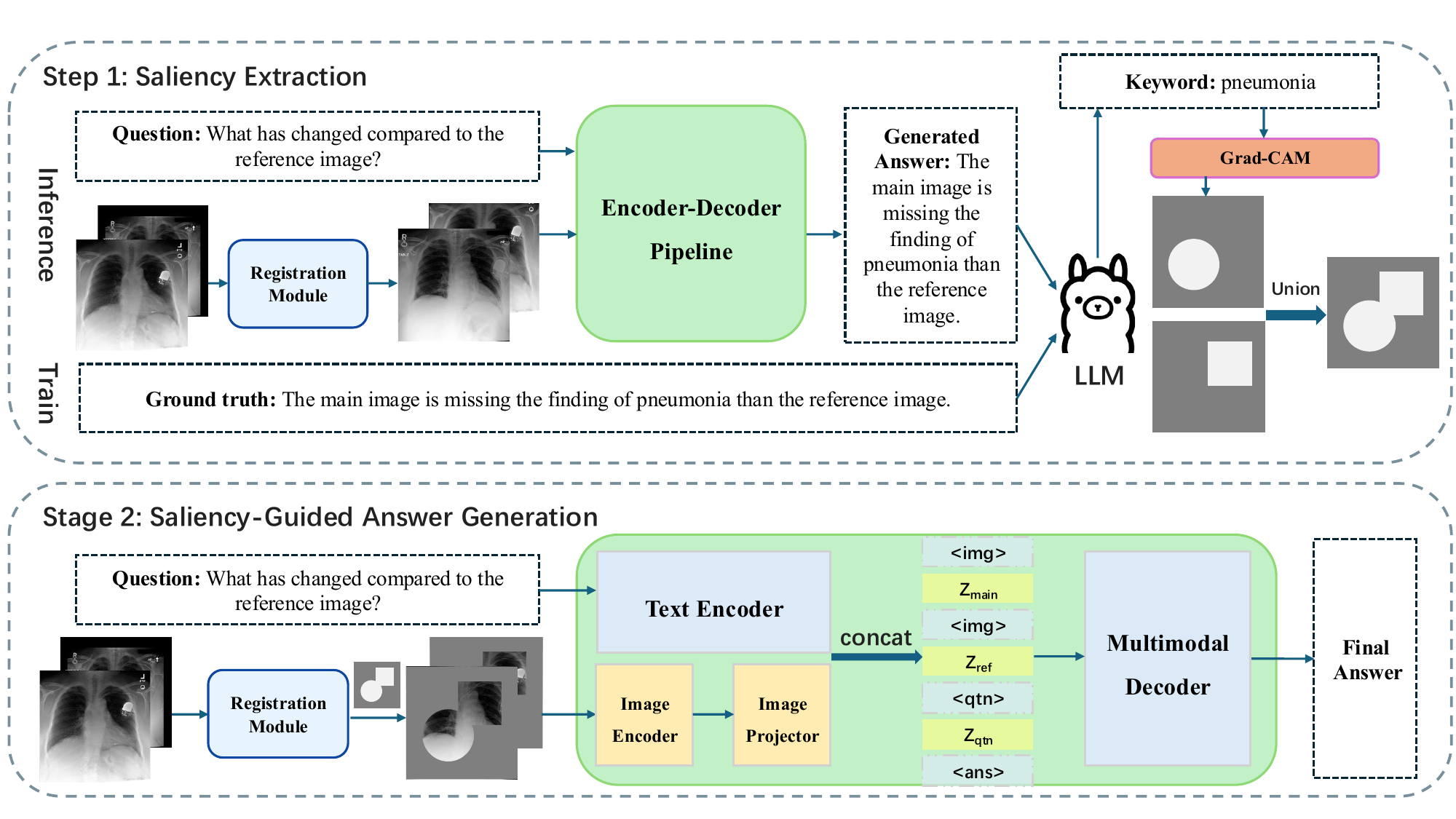}
  \caption{Overview of our proposed method. The step 1 is for saliency extraction, and step 2 is for answer generation. Saliency extraction and application are based on the after-registration images. Grad-CAM uses the keyword as the explanation target. $Z_{\text{main}}$, $Z_{\text{ref}}$, $Z_{\text{qtn}}$ are extracted representations for the main image, reference image and question. \(\langle\cdot\rangle\) denotes special tokens for separation.}
  \label{fig:arch}
\end{figure}

%\subsection{Architecture}
An overview appears in Figure~\ref{fig:arch}. The pipeline has two components: a micro image registration module and a keyword-conditioned saliency extraction module, followed by image–text encoders and a multimodal decoder.

\subsection{Micro Image Registration Module}
Given a main image \(I_{\text{main}}\in\mathbb{R}^{3\times H\times W}\) and a reference image \(I_{\text{ref}}\in\mathbb{R}^{3\times H\times W}\), a shallow CNN predicts 2D affine parameters \(\Theta=[A\;\mathbf{t}]\in\mathbb{R}^{2\times 3}\). We warp only the main image with a differentiable grid sampler:
\[
\mathbf{x} = A\,\mathbf{x}_{\mathrm{tgt}} + \mathbf{t}.
\]
To keep the transform near identity and avoid erasing true anatomical change, we regularize
\[
\mathcal{L}_{\text{reg}} = w_{\text{small}}\|\Theta - I\|_F^{2} + w_{\det}(\det(A)-1)^{2} + w_{\text{trans}}\|\mathbf{t}\|_2^{2},
\]
with \(w_{\text{small}}=10^{-4}\), \(w_{\det}=10^{-5}\), and \(w_{\text{trans}}=10^{-6}\). The registered image is \(\widehat{I}_{\text{main}}\).

\subsection{Saliency Extraction Module}
We compute gradient-based saliency on the registered pair using Grad-CAM~\citep{grad-cam, jacobgilpytorchcam}. A single clinical keyword is extracted from the answer using Llama 3:70B~\citep{llama3modelcard, OllamaOllamaGet} and used as the target concept for saliency. We obtain maps \(S_{\text{main}}\) and \(S_{\text{ref}}\) on \(\widehat{I}_{\text{main}}\) and \(I_{\text{ref}}\), respectively, then form a shared mask by element-wise maximum, $S = \max(S_{\text{main}},\, S_{\text{ref}}).$
After min–max normalization of \(S\) to \([0,1]\), we apply it multiplicatively to both images:
\[
I'_{\text{main}} = S \odot \widehat{I}_{\text{main}}, \qquad I'_{\text{ref}} = S \odot I_{\text{ref}}.
\]
This encourages consistent focus on corresponding anatomy at both time points while retaining non-salient context with attenuated weight.

\subsection{Encoders and Decoders}
\paragraph{Image encoder and projector.}
We use ResNet-50~\citep{resnet} pretrained on ImageNet-1k~\citep{imagenet_cvpr09}. From its penultimate feature map \(\mathbb{R}^{H\times W\times C}\) we form a token sequence \(\mathbb{R}^{N\times C}\) with \(N=HW\). A projector with one linear layer, one 8-head Transformer encoder~\citep{transformer}, and a two-layer MLP maps image tokens to the text-representational space.

\paragraph{Text encoder.}
Questions are tokenized with embeddings shared by the decoder, then added with a learnable positional embedding and passed through six 12-head Transformer encoder layers.

\paragraph{Multimodal decoder.}
A GPT-2~\citep{gpt2} decoder from HuggingFace~\citep{hf} consumes masked image tokens and question tokens to generate the answer. We add special tokens \(\langle\text{pad}\rangle\), \(\langle\text{img}\rangle\), \(\langle\text{qtn}\rangle\), and \(\langle\text{ans}\rangle\). Denote the main image representation as $Z_{\text{main}}$, the reference image representation as $Z_{\text{ref}}$, and the question representation as $Z_{\text{qtn}}$ .The input sequence is
\[
\textbf{concat}(\langle\text{img}\rangle, Z_{\text{main}}, \langle\text{img}\rangle, Z_{\text{ref}}, \langle\text{qtn}\rangle, Z_{\text{qtn}}, \langle\text{ans}\rangle).
\]

\subsection{Training and Inference}
\paragraph{Training.}
During training, the ground-truth answer provides the keyword for saliency. The total loss is the sum of registration and language modeling terms, $\mathcal{L}_{\text{Total}}=\mathcal{L}_{\text{reg}}+\mathcal{L}_{\text{LM}}$. We run a 1-epoch warm-up without saliency masking, then enable the two-step loop for the remaining epochs of a 16-epoch schedule.

% (brief introduction to metrics used, including their semantic meaning/function; also discuss the "ourscore" in the metric.py; remember to give the citations.)

\paragraph{Inference.}
We use a two-pass procedure. First, the decoder generates a preliminary answer without masking, from which we extract a keyword. Second, we compute keyword-conditioned saliency on both images, apply the shared mask, and regenerate the final answer using the masked inputs. For latency-sensitive use, a single-pass variant without masking is available, but the two-pass variant better enforces longitudinal consistency.

\section{Results}

We adopt common generation metrics in VQA, BLEU-1~\citep{papineni-etal-2002-bleu} (n-gram precision with a brevity penalty), METEOR~\citep{banerjee-lavie-2005-meteor} (stem matching with an emphasis on recall), ROUGE-L~\citep{lin-2004-rouge} (overlap and longest common subsequence), and CIDEr~\citep{vedantam2015ciderconsensusbasedimagedescription} (a TF-IDF based consensus metric) to evaluate different aspects such as surface-level matching, semantic alignment, and consistency with human references. To emphasize the medical keyword and semantic meaning, also considering the scale of the metrics, we combine METEOR and CIDEr as
\[
0.6 \cdot \frac{\mathrm{CIDEr}}{1 + \mathrm{CIDEr}} + 0.4 \cdot \mathrm{METEOR},
\]
to select the model that performs the best at the end of training.

\begin{table}[!htbp]
  \caption{Evaluation on Medical-Diff-VQA comparing ours with prior works. 
  The ``Pre-train (med.)'' column indicates whether a method uses additional pre-training on medical data (``Yes'') or not (``No'').}
  \vspace{\baselineskip} 
  \label{tab:benchmark}
  \centering
  \resizebox{0.7\textwidth}{!}{%
  \begin{tabular}{lccccc}
    \toprule
    Methods & Pre-train (med.) & BLEU-1 & METEOR & ROUGE-L & CIDEr \\
    \midrule
    MCCFormers~\citep{mccformers} & \textbf{No}  & 0.214 & 0.319 & 0.340 & 0     \\
    IDCPCL~\citep{idcpcl,reai}    & \textbf{No}  & 0.614 & 0.303 & 0.582 & 0.703 \\\hline
    EKAID~\citep{ekaid,reai}      & Yes & 0.628 & 0.339 & 0.557 & 1.027 \\
    RegioMix~\citep{regio}        & Yes & 0.705 & 0.381 & 0.651 & 1.804 \\
    PLURAL~\citep{plural}         & Yes & 0.704 & 0.381 & 0.653 & 1.832 \\
    ReAl~\citep{reai}             & Yes & 0.710 & 0.395 & 0.736 & 2.409 \\
    VED~\citep{ved}               & Yes & 0.716 & 0.389 & 0.670 & 2.119 \\
    \midrule
    Ours                          & \textbf{No}  & 0.628 & 0.651 & 0.627 & 1.263 \\
    \bottomrule
  \end{tabular}
  }
\end{table}

% % ; best scores are in \textbf{bold}
% \begin{table}[!htbp]
%   \caption{Evaluation on Medical-Diff-VQA comparing ours with prior works.}
%   \vspace{\baselineskip} 
%   \label{tab:benchmark}
%   \centering
%   \resizebox{1\textwidth}{!}{%
%   \begin{tabular}{lccccccc}
%     \toprule
%     Methods & BLEU-1 & BLEU-2 & BLEU-3 & BLEU-4 & METEOR & ROUGE-L & CIDEr \\
%     \midrule
%     MCCFormers~\citep{mccformers} & 0.214 & 0.190 & 0.170 & 0.153 & 0.319 & 0.340 & 0     \\
%     IDCPCL~\citep{idcpcl,reai}     & 0.614 & 0.541 & 0.474 & 0.414 & 0.303 & 0.582 & 0.703 \\
%     EKAID~\citep{ekaid,reai}       & 0.628 & 0.553 & 0.491 & 0.434 & 0.339 & 0.557 & 1.027 \\
%     RegioMix~\citep{regio}   & 0.705 & 0.633 & 0.572 & 0.517 & 0.381 & 0.651 & 1.804 \\
%     PLURAL~\citep{plural}     & 0.704 & 0.633 & 0.575 & 0.520 & 0.381 & 0.653 & 1.832 \\
%     ReAl~\citep{reai}  & 0.710 & 0.636 & 0.580 & 0.530 & 0.395 & 0.736 & 2.409 \\
%     VED~\citep{ved} & 0.716 & 0.647 & 0.590 & 0.537 & 0.389 & 0.670 & 2.119 \\
%     \midrule
%     Ours & 0.628 & 0.510 & 0.418 & 0.341 & 0.651 & 0.627 & 1.263 \\
%     \bottomrule
%   \end{tabular}
%   }
% \end{table}

Table \ref{tab:benchmark} shows that, although our BLEU-1 scores are modest, they still indicate non-trivial lexical precision: the model reliably reproduces key clinical tokens (e.g., anatomy, laterality, lesion attributes) across paired studies rather than collapsing to generic templates. Conversely, METEOR is notably high (0.651), evidencing strong synonym and inflectional coverage, while ROUGE-L (0.627) and CIDEr (1.263) suggest that the generated answers preserve sentence-level coverage and place appropriate emphasis on TF-IDF-informative, clinically salient phrases. Together, these metrics indicate that saliency guidance and keyword-conditioned targets help the decoder to provide a focus on disease-bearing regions/terms, resulting in good semantic adequacy with more conservative n-gram precision.

The overall room for improvement is understandable. First, none of our components were adapted to medical data: the general-purpose image backbone and text decoder were used off-the-shelf without further pretraining on radiology images and texts, and the keyword extractor relied on a general-purpose LLM rather than a domain-specialized model. Second, the current checkpoint was trained for 16 epochs, which likely limited the convergence of our model; longer training (with an extended warm-up before enabling saliency) could benefit n-gram precision. Third, our registration is intentionally lightweight and near-identity, which may under-correct inter-visit motion differences; such residual misalignment may depress BLEU-n while still allowing METEOR/CIDEr to reflect correct semantics.

% If results not good, illustrate that we are using basic models.
% Table ~\ref{tab:benchmark} reveals that our method achieves good performance in BLEU-1/2/3/4 (measures accurate reproduction of key terms such as anatomical locations and lesion attributes) and METEOR (sensitivity to synonyms and inflectional forms)). VED showed suboptimal performance across the BLEU series (e.g., BLEU-4=0.537), while EIE-all achieved the best METEOR score among existing methods (0.401). In contrast, ReAl achieved the best performance on ROUGE-L (sentence-level coverage and sequence consistency based on longest common subsequences) and CIDEr (TF-IDF-weighted multi-reference consistency, emphasizing clinically rare yet critical information such as “pneumothorax” and " loculated effusion"). Our method ranks second (0.70, 2.20). In summary, the results demonstrate that our method exhibits significant advantages in lexical-level precision and semantic-level adequacy, while maintaining strong competitiveness in paragraph-level coverage and group consensus, second only to ReAl. This reflects a more balanced improvement across different evaluation dimensions that aligns with the characteristics of medical texts.

\section{Acknowledgment}
This work was partially supported by NIH R21EB034911.
\bibliographystyle{unsrt}
\bibliography{refs}

%%%%%%%%%%%%%%%%%%%%%%%%%%%%%%%%%%%%%%%%%%%%%%%%%%%%%%%%%%%%
\newpage
\appendix
\section{Appendix and Discussions}
\label{sec:conclusion}
This paper proposes a generative framework for longitudinal medical image differential question-answering. It combines near-identity affine registration, dual-image encoding, and a text-image generation decoder to establish a closed-loop process from keywords to saliency (CAM) to feature weighting. This ensures consistent alignment between linguistic cues and visual evidence during both training and inference phases, thereby enhancing the localizability of lesion-related regions and the interpretability of answers. It is crucial to emphasize that all backbone models and decoders in this work are pre-trained on general-purpose datasets without further training on medical data. The LLM used for keyword extraction is also a general-purpose model, further highlighting the method's transferability and engineering feasibility.

Despite the aforementioned progress, this study has limitations. The additional saliency extraction step introudces large time complexity. When keyword extraction relies on general-purpose LLMs rather than medically specialized pre-trained models, it may omit critical lesion terminology or introduce overly generic vocabulary, thereby weakening visual alignment effectiveness. Additionally, the currently employed affine registration is relatively simple and may struggle to accommodate drastic variations in pose and imaging conditions. Apart from that, since current data and tasks focus on the difference problem within the MIMIC framework, cross-dataset generalization remains to be validated. Furthermore, the absence of medical retraining may limit the upper performance bound. Finally, as this is an ongoing study, further exploration is needed regarding the impact of additional ablation studies, keyword extraction and saliency method selection, and larger encoders on model performance.

Despite these limitations, this study holds substantial significance and value. We pioneered the conversion of medically relevant keywords automatically extracted by general-purpose LLMs into saliency targets for feature weighting. This establishes a closed-loop mechanism where language supervision directly constrains visual attention, systematically enhancing evidence utilization and model interpretability without requiring medical pre-training. Given that medical question-answering is often driven by lesion nouns and key anatomical locations, these keywords provide sparse yet strongly constrained supervisory signals, demonstrating both necessity and novelty. The comprehensive workflow provides a reusable baseline and engineering paradigm for future extensions to stronger visual backbones, more sophisticated deformation models, and larger language models. 

% The comprehensive data processing workflow, registration-encoding-decoding architecture, and fusion scoring strategy provide a reusable baseline and engineering paradigm for future extensions to stronger visual backbones, more sophisticated deformation models, and larger language models. 
% This approach demonstrates robust gains and strong transferability even when using only general pre-trained models.

\end{document}